\documentclass[11pt]{article}

\usepackage{acl}
\usepackage{multirow}
\usepackage{times}
\usepackage{latexsym}
\usepackage[T1]{fontenc}
\usepackage[utf8]{inputenc}
\usepackage{microtype}
\usepackage{inconsolata}
\usepackage{graphicx}
\usepackage{amsmath}
\usepackage{amssymb}
\usepackage{booktabs}
\usepackage{algorithm}
\usepackage{algpseudocode}

\newcommand{\cohyde}{CoHyDE}

\title{\cohyde{}: Iterative Co-Training of LLM Rewriter \&\\Dense Encoder for Tool Retrieval}

\author{
  Vaishali Senthil\thanks{Equal contribution.} \qquad
  Ashutosh Hathidara\footnotemark[1] \qquad
  Sebastian Schreiber \\
  SAP Labs \\
  {\small \texttt{\{vaishali.senthil, ashutosh.hathidara, sebastian.schreiber\}@sap.com}}
}

\begin{document}
\maketitle

\begin{abstract}
Tool retrieval over large API catalogs is a core bottleneck for LLM agents: user queries arrive in colloquial, often underspecified language, while the catalog uses technical API vocabulary that no fixed encoder can bridge on its own.
The two dominant training approaches, contrastive encoder fine-tuning and HyDE-style query expansion with a frozen LLM, address this problem from opposite ends and fail in complementary directions: the fine-tuned encoder excels when the query's surface form already matches the catalog but collapses when it does not, while zero-shot HyDE is more robust to underspecified queries yet generates catalog-unaware hypothetical descriptions that degrade retrieval when queries are well-formed.
We introduce \textbf{\cohyde{}}, an iterative procedure that trains the dense encoder and the LLM rewriter as a single co-evolving system: the encoder is retrained with InfoNCE on catalog-style hypothetical descriptions produced by the rewriter, and the rewriter is preference-aligned via DPO against the encoder's retrieval scores, with both sides warm-started on the tool catalog before the loop begins.
On a $\sim$10k tool subset of the ToolBench catalog~\citep{qin2024toolllm}, three rounds of \cohyde{} improve over the strongest single-component baseline by \textbf{+2.5~pp} NDCG@5 on standard queries and \textbf{+6.3~pp} on held-out vague queries, with gains as large as \textbf{+8~pp} on cross-domain vague queries~(G2).
Ablations confirm that co-training is the key ingredient: using either component in isolation fails to match \cohyde{} on both well-formed and vague queries, with losses of up to \textbf{-8~pp} on vague queries.
\end{abstract}

\section{Introduction}
\label{sec:intro}

Modern language model agents act in the world by calling external tools drawn from catalogs that increasingly number in the tens of thousands \citep{qin2024toolllm,patil2023gorilla}. No agent can fit every tool's documentation into its context window, and the quality of an agent's actions is bounded above by an upstream \emph{tool retrieval} step that selects a small candidate set per user query.

The dominant retrieval recipe embeds queries and tools into a shared vector space and returns the top-$k$ most similar tools by nearest-neighbor lookup. Two largely disjoint research directions have grown around this recipe. \textbf{Direction 1: query expansion with a frozen LLM.} HyDE-style methods \citep{gao2023hyde,wang2023query2doc} prompt a frozen LLM to generate a hypothetical document for the query and search a frozen encoder against its embedding. \textbf{Direction 2: encoder fine-tuning with no query rewriting.} Dense-retrieval methods fine-tune the encoder on (query, tool) pairs with contrastive losses \citep{karpukhin2020dpr,xiao2023bge}.

Both directions have a complementary failure mode. A trained dense encoder is, in essence, a similarity function shaped by the (anchor, positive) pairs it sees during training. When the query is in-distribution (i.e., sharing lexical surface with the catalog), the contrastive signal is sufficient; when surface form drifts, the encoder has no world-knowledge or reasoning machinery to bridge the gap and falls back on residual lexical cues \citep{thakur2021beir,chen2022spar}. Query-expansion approaches fail symmetrically: the LLM brings the reasoning needed to handle vague queries \citep{wei2022cot}, but its generated output does not match the catalog's vocabulary, so on well-formed queries, it hurts more than it helps \citep{lei2024csqe}. This raises a natural question: \emph{can the two training modes be combined into a single procedure that is stronger than either component alone?}

We introduce \textbf{\cohyde{}}, an iterative co-training procedure that treats the dense encoder and the LLM rewriter as a single co-evolving system. In each round, the LLM generates catalog-style hypothetical descriptions for each query; the encoder is then retrained via contrastive learning on these descriptions, and the LLM is preference-aligned via DPO using the encoder's own retrieval scores as reward signal. This alternating update cycle is repeated for multiple iterations, with each component progressively adapting to the other.

We apply \cohyde{} on a $\sim$10k-tool subset of the ToolBench catalog~\citep{qin2024toolllm}. After three rounds of co-training, \cohyde{} improves over the strongest single-component baseline by \textbf{+2.5~pp} NDCG@5 on standard queries and \textbf{+6.3~pp} on held-out vague queries, with gains as large as \textbf{+8~pp} on cross-domain vague queries~(G2).

To summarize our contributions:
(i) We introduce \cohyde{}, an iterative co-training procedure that jointly optimizes a dense encoder and an LLM rewriter for tool retrieval.
(ii) We empirically characterize the complementary failure modes of encoder fine-tuning and zero-shot HyDE, motivating the need to train both components jointly.

\section{Related Work}
\label{sec:related}

\paragraph{Tool retrieval.} Dense tool-retrieval methods fine-tune an encoder on (query, tool) pairs with contrastive supervision \citep{qin2024toolllm,anantha2023protip,qu2024colt,shi2025toolret}; a parallel line treats retrieval as a frozen black-box via LLM-based expansion or generative indexing \citep{patil2023gorilla,chen2024reinvoke,lumer2024toolshed,wang2024toolgen,toolsense2026}. The closest prior work is \citet{xu2024iterative}, which iteratively rewrites user instructions and retrains the encoder on (rewritten-instruction, tool) pairs. \cohyde{} differs: the rewriter is preference-aligned via DPO against the encoder it feeds, rewrites target \emph{catalog-description style} rather than query style, and the encoder retrain uses no real (query, tool) pairs. 

\paragraph{Query expansion and trained rewriters.} HyDE \citep{gao2023hyde} searches a frozen index against a hypothetical document embedding; Query2doc \citep{wang2023query2doc} concatenates the pseudo-document to the original query. CSQE \citep{lei2024csqe} patches corpus-misalignment of LLM expansions at test time by injecting retrieved sentences; we address the same misalignment at training time. Trained query rewriters like Rewrite-Retrieve-Read \citep{ma2023rewrite}, RaFe \citep{mao2024rafe}, and LeReT \citep{hsu2024leret} use RL or DPO with a \emph{frozen} retriever; a complementary thread \citep{nogueira2019doc2query,dai2022promptagator,bonifacio2022inpars,wang2022gpl} trains the retriever on LLM-generated synthetic queries with the generator frozen. All these methods freeze at least one component, whereas \cohyde{} co-trains both.

\paragraph{Dense retriever robustness and joint retriever-generator training.} Dense retrievers are brittle off-distribution \citep{thakur2021beir,sciavolino2021entityquestions,chen2022spar,yu2022cocodr}; domain-adaptation via synthetic queries \citep{wang2022gpl,dai2022promptagator,meng2024augtriever,lin2023dragon} runs the generation loop once with a frozen generator. Joint retriever--generator frameworks like RAG \citep{lewis2020rag}, Atlas \citep{izacard2023atlas}, REPLUG \citep{shi2024replug}, RA-DIT \citep{lin2024radit}, Self-RAG \citep{asai2024selfrag} train the generator to produce better \emph{final answers}, not better retrieval inputs. Prior work has therefore never co-trained a generator whose output \emph{is} the retrieval input with the encoder that consumes it, the precise gap \cohyde{} fills.

\section{Methodology}
\label{sec:method}

\begin{figure*}[t]
  \centering
  \includegraphics[width=\textwidth]{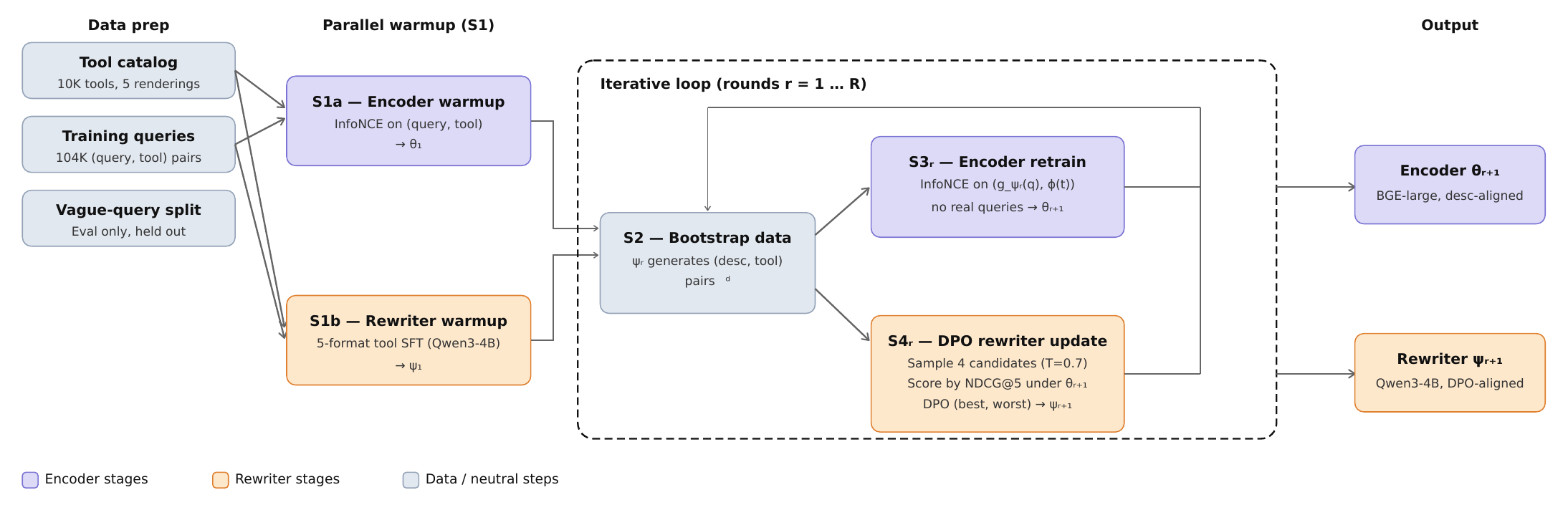}
  \caption{Overview of \cohyde{}: a dense encoder and an LLM rewriter are co-trained in an alternating loop, with each component iteratively adapted to the other.}
  \label{fig:methodology}
\end{figure*}

\subsection{Problem Formulation}

Let $\mathcal{T} = \{t_{1}, \ldots, t_{N}\}$ denote a tool catalog of size $N$, where each tool $t \in \mathcal{T}$ carries a structured record (api name \& description as well as tool title \& description). We write $\phi : \mathcal{T} \to \Sigma^{*}$ for a fixed \emph{rendering} function that serialises a tool into a single text string. Given a query $q \in \Sigma^{*}$ and a budget $k \in \mathbb{N}$, the tool-retrieval problem is to return a ranked set $\hat{T}_{k}(q) \subseteq \mathcal{T}$ with $|\hat{T}_{k}(q)| = k$ that maximally overlaps the gold tool set $T^{*}_{q} \subseteq \mathcal{T}$.

We restrict attention to single-vector dense encoder retrieval, the dominant architecture in tool retrieval \citep{qin2024toolllm,anantha2023protip,qu2024colt,shi2025toolret}. A parameterised encoder $f_{\theta} : \Sigma^{*} \to \mathbb{R}^{d}$ maps any text into a $d$-dimensional unit-norm vector and retrieval is performed by approximate nearest-neighbour search \citep{johnson2019faiss},
\begin{equation}
\hat{T}_{k}(q;\theta) = \mathrm{topk}_{t \in \mathcal{T}}\, \bigl\langle f_{\theta}(q),\; f_{\theta}\!\bigl(\phi(t)\bigr) \bigr\rangle
\end{equation}
We additionally consider a \emph{rewriter-augmented} variant in which a generator $g_{\psi} : \Sigma^{*} \to \Sigma^{*}$ produces a hypothetical tool description $\tilde{d} = g_{\psi}(q)$ that is encoded \emph{in place of} the query:
\begin{equation}
\hat{T}_{k}^{g_{\psi}}(q;\theta) = \mathrm{topk}_{t \in \mathcal{T}}\, \bigl\langle f_{\theta}\!\bigl(g_{\psi}(q)\bigr),\; f_{\theta}\!\bigl(\phi(t)\bigr) \bigr\rangle.
\end{equation}
The goal of \cohyde{} is to find parameters $(\theta^*, \psi^*)$ such that the two components reinforce each other, which we achieve through an alternating sequence of encoder and rewriter updates described in \S\ref{sec:cotrain}.

\subsection{Data}
\label{subsec:data-in-method}

\paragraph{Tool catalog.} The ToolBench API pool \citep{qin2024toolllm} contains $|\mathcal{T}_{\mathrm{full}}| = 46{,}980$ tools, partitioned into three evaluation tiers: single-domain (G1), cross-domain same-category (G2), and cross-domain different-category (G3), with 1{,}092 official evaluation queries (593 / 399 / 100 over G1/G2/G3). We work with a stratified subset $\mathcal{T}$ of $N = 10{,}000$ tools sized for training tractability: the subset retains every tool referenced by the gold sets of the evaluation queries, and stratified-samples the remaining slots to preserve the per-tier proportions of $\mathcal{T}_{\mathrm{full}}$.

\paragraph{Training set.} The training set $\mathcal{D}_{\mathrm{train}} = \{(q_{i}, T^{*}_{q_{i}})\}_{i=1}^{M}$ consists of $M = 104{,}224$ (query, gold-tool-set) pairs (44{,}873 / 35{,}402 / 23{,}949 over G1/G2/G3); most queries have multiple gold tools ($|T^{*}_{q}| > 1$ for 93--99\% of $q$). For contrastive training, we flatten these to individual (query, tool) pairs $\mathcal{D}_{\mathrm{q}} = \{(q, \phi(t)) : (q, T^{*}_{q}) \in \mathcal{D}_{\mathrm{train}},\, t \in T^{*}_{q}\}$.
A fixed 10\% of $\mathcal{D}_{\mathrm{train}}$ is held out as a \emph{dev split} for checkpoint selection at S1a and each S3$_{r}$, and for selecting the number of co-training rounds $R$ reported in \S\ref{sec:ablations} (stratified by tier, seed~$=42$; 4{,}487\,/\,3{,}540\,/\,2{,}395 queries over G1/G2/G3); it is fully disjoint from the 1{,}092-query evaluation set and $\mathcal{Q}_{\mathrm{vague}}$, and no evaluation query or its gold tools are seen during any training stage.

\paragraph{Tool rendering.} We represent each tool under a family of five rendering conventions $\Phi = \{\phi_{1}, \ldots, \phi_{5}\}$ spanning its natural information axes: $\phi_{1}$ (title only), $\phi_{2}$ (+API name), $\phi_{3}$ (+tool description), $\phi_{4}$ (title, API name, API description), and $\phi_{5}$ (full record). At training time, $\phi \sim \mathrm{Unif}(\Phi)$ is sampled independently per (query, tool) pair, so each tool is seen under all five surface forms over an epoch. This format mixture encourages the encoder to learn representations invariant to catalog-side surface variation, including the longer multi-sentence $\phi_{5}$ that most closely matches the rewriter's output style. At inference, the catalog is indexed under $\phi_{5}$.

\paragraph{Vague-query split.} We adopt the vague-query evaluation protocol of \citet{chen2026trb} to probe robustness under query-side distribution shift. Each $q \in \mathcal{Q}_{\mathrm{eval}}$ is paraphrased to replace surface tokens with conversational alternatives, while preserving the original gold tool set. We follow the protocol of \citet{chen2026trb} exactly, substituting claude-4.5-opus for the GPT-4o paraphraser used in the original work. $\mathcal{Q}_{\mathrm{vague}}$ does not enter any training procedure; two-pass validation (LLM self-check on every paraphrase plus an author spot-check on 50 samples) is described in Appendix~\ref{app:vague}.

\subsection{Encoder}

$f_{\theta}$ is initialised from BGE-large-en-v1.5 \citep{xiao2023bge} ($\approx$335M parameters, $d=1024$). Given $x \in \Sigma^{*}$ we define $f_{\theta}(x) = h^{\theta}_{\mathrm{CLS}}(x) / \| h^{\theta}_{\mathrm{CLS}}(x) \|_{2}$, and the same encoder is applied to queries, tool renderings, and rewriter outputs (a \emph{symmetric} bi-encoder). Training minimises the symmetric InfoNCE loss \citep{oord2018cpc} with temperature $\tau = 0.05$ and in-batch negatives; full loss expression and optimisation hyperparameters are in Appendix~\ref{app:encoder-hp}.

We define two contrastive training datasets, differing only in what serves as the anchor: $\mathcal{D}_{\mathrm{q}}$ pairs user queries with tool renderings, while $\mathcal{D}^{(\psi)}_{\mathrm{d}}$ pairs rewriter-generated hypothetical descriptions $g_{\psi}(q)$ with tool renderings. In both cases the tool side is rendered under a rendering $\phi$ sampled uniformly from $\Phi$.

\subsection{Rewriter}
\label{sec:rewriter}

$g_{\psi}$ is Qwen3.5-4B \citep{qwen3report}, an instruction-tuned decoder-only transformer. We define a prompt operator $\rho_{\mathrm{HyDE}}$ that wraps a query with an instruction to enumerate the full tool description of tool capable of fulfilling the query's intent, in catalog-style description format (Appendix~\ref{app:hyde-prompt}). A deterministic cleaning operator $\mathrm{clean}(\cdot)$ strips reasoning-trace blocks and conversational preambles before encoding (Appendix~\ref{app:clean}). At inference, the rewriter produces $\tilde{d} = \mathrm{clean}(g_{\psi}(\rho_{\mathrm{HyDE}}(q)))$ and retrieval proceeds against $\tilde{d}$ alone, replacing the original query entirely.

\subsection{\cohyde{}: Iterative Co-training}
\label{sec:cotrain}

We index encoder and rewriter checkpoints by training stage: $\theta_{i}$, $\psi_{i}$ are the parameters after stage $i$. $\theta_{0}$ denotes BGE-large-en-v1.5 pretrained weights; $\psi_{0}$ denotes the opensource instruction-tuned Qwen3.5-4B. The pipeline has two parallel warmup steps (S1a \& S1b) followed by a bootstrap data-generation step (S2) and an alternating training loop (S3, S4) that may be unrolled for any number of rounds $R \geq 1$. Figure~\ref{fig:methodology} and Algorithm~\ref{alg:cotrain} summarise the procedure.

\begin{algorithm}[t]
\caption{\cohyde{}: Iterative Co-Training}
\label{alg:cotrain}
\begin{algorithmic}[1]
\Require Pretrained encoder $\theta_{0}$, base rewriter $\psi_{0}$, training pairs $\mathcal{D}_{\mathrm{train}}$, prompt $\rho_{\mathrm{HyDE}}$, rendering family $\Phi$, rounds $R$
\Ensure Co-trained encoder $\theta_{R+1}$ and rewriter $\psi_{R+1}$

\State \textbf{[S1a]} \textit{Encoder warmup:} train $\theta_{0}$ with InfoNCE on $\{(q,\, \phi_{5}(t))\}$ from $\mathcal{D}_{\mathrm{train}}$ to obtain $\theta_{1}$
\State \textbf{[S1b]} \textit{Rewriter warmup:} fine-tune $\psi_{0}$ on catalog tools under $\Phi$ to obtain $\psi_{1}$
\State \textbf{[S2]} \textit{Bootstrap:} generate $\mathcal{D}^{(\psi_{1})}_{\mathrm{d}} = \{(g_{\psi_{1}}(\rho_{\mathrm{HyDE}}(q)),\, \phi(t))\}$ for $(q,t) \in \mathcal{D}_{\mathrm{train}}$

\For{$r = 1, \ldots, R$}
    \State \textbf{[S3$_r$]} \textit{Encoder retrain:} train $\theta_{r}$ with InfoNCE on $\mathcal{D}^{(\psi_{r})}_{\mathrm{d}}$ to obtain $\theta_{r+1}$
    \State \textbf{[S4$_r$]} \textit{Rewriter alignment:}
    \State \quad Sample $N$ descriptions $\{\tilde{d}^{(j)}\} \sim g_{\psi_{r}}(\rho_{\mathrm{HyDE}}(q))$ for each $q \in \mathcal{D}_{\mathrm{train}}$
    \State \quad Score each $\tilde{d}^{(j)}$ by NDCG@5 under $\theta_{r+1}$
    \State \quad Form preference pair: $\tilde{d}^{+}_{q} = \arg\max_j\,\mathrm{NDCG@5}(\tilde{d}^{(j)})$,\; $\tilde{d}^{-}_{q} = \arg\min_j$
    \State \quad $\psi_{r+1} \gets \arg\min_{\psi}\, \mathcal{L}_{\mathrm{DPO}}(\psi;\, \psi_{r})$
    \State \quad $\mathcal{D}^{(\psi_{r+1})}_{\mathrm{d}} = \{(g_{\psi_{r+1}}(\rho_{\mathrm{HyDE}}(q)),\, \phi(t))\}$
\EndFor
\State \Return $(\theta_{R+1},\, \psi_{R+1})$
\end{algorithmic}
\end{algorithm}

\paragraph{S1a: Encoder warmup.} The encoder is trained with InfoNCE on (query, tool) pairs from $\mathcal{D}_{\mathrm{train}}$:
\begin{equation}
\theta_{1} = \arg\min_{\theta}\, \mathbb{E}_{(q,t)\sim\mathcal{D}_{\mathrm{train}}}\, \mathcal{L}_{\mathrm{NCE}}\bigl(\theta; (q, \phi_{5}(t))\bigr)
\end{equation}
This is the standard contrastive tool-retrieval recipe \citep{qin2024toolllm,anantha2023protip,shi2025toolret}, and is observed to be the strongest encoder-only baseline (Table~\ref{tab:main}). We initialise the loop from $\theta_{1}$ rather than pretrained BGE so the encoder has a contrastive head start before description-only retraining begins.

\paragraph{S1b: Rewriter warmup.} The rewriter is fine-tuned on the catalog itself, with each tool $t$ shown under all five renderings $\phi_{1}, \ldots, \phi_{5}$ from the format family $\Phi$ (defined in \S\ref{subsec:data-in-method}):
\begin{equation}
\psi_{1} = \arg\min_{\psi}\, -\!\!\sum_{t \in \mathcal{T}}\sum_{\phi_{i} \in \Phi} \log p_{\psi}\bigl(\phi_{i}(t)\bigr)
\end{equation}
This teaches the rewriter the catalog's vocabulary, naming conventions, and the multiple surface forms a tool can take.

\paragraph{S2: Bootstrap data generation.} Using $\psi_{1}$ and the prompt $\rho_{\mathrm{HyDE}}$, we generate the first round of (description, tool) training data:
\begin{equation}
\mathcal{D}^{(\psi_{1})}_{\mathrm{d}} = \bigl\{(g_{\psi_{1}}(\rho_{\mathrm{HyDE}}(q)), \phi(t)) : (q, t) \in \mathcal{D}_{\mathrm{train}}\bigr\}
\end{equation}
with $\phi \sim \mathrm{Unif}(\Phi)$. The 5-format-trained rewriter produces catalog-style tool descriptions, used as the contrastive anchors for the next encoder training.

\paragraph{S3$_{r}$: Encoder retraining.} For each round $r = 1, \ldots, R$, the encoder is trained further on $\mathcal{D}^{(\psi_{r})}_{\mathrm{d}}$, continuing from $\theta_{r}$:
\begin{equation}
\theta_{r+1} = \arg\min_{\theta}\, \mathbb{E}\, \mathcal{L}_{\mathrm{NCE}}\bigl(\theta; (g_{\psi_{r}}(\rho_{\mathrm{HyDE}}(q)), \phi(t))\bigr)
\end{equation}
No real $(q, t)$ pair participates in this stage; the encoder is trained \emph{only} on $(g_{\psi_{r}}(q), \phi(t))$ pairs.

\paragraph{S4$_{r}$: DPO alignment of the rewriter.} For each $q \in \mathcal{D}_{\mathrm{train}}$, sample $N$ candidate descriptions $\{\tilde{d}^{(j)}\} \sim g_{\psi_{r}}(\rho_{\mathrm{HyDE}}(q))$ and score them by NDCG@5 under the just-trained encoder $\theta_{r+1}$. Form a preference pair $(\tilde{d}^{+}_{q}, \tilde{d}^{-}_{q})$ from the argmax and argmin of those scores, and minimise the standard DPO objective \citep{rafailov2023dpo}:
\begin{multline}
\mathcal{L}_{\mathrm{DPO}}(\psi;\psi_{r}) = -\mathbb{E}_{q}\,\log\sigma\Biggl(\beta\log\frac{p_{\psi}(\tilde{d}^{+}_{q} | \rho(q))}{p_{\psi_{r}}(\tilde{d}^{+}_{q} | \rho(q))} \\
- \beta\log\frac{p_{\psi}(\tilde{d}^{-}_{q} | \rho(q))}{p_{\psi_{r}}(\tilde{d}^{-}_{q} | \rho(q))}\Biggr)
\end{multline}
$\psi_{r+1} = \arg\min_{\psi}\, \mathcal{L}_{\mathrm{DPO}}(\psi;\psi_{r})$ is then used to regenerate $\mathcal{D}^{(\psi_{r+1})}_{\mathrm{d}}$ for the next round. The encoder of round $r$ supervises the rewriter update, and the rewriter of round $r+1$ produces the data for the next encoder update; both sides evolve along a coupled trajectory.

\subsection{Evaluation Protocol}
\label{sec:eval}

We report hit@$k$, recall@$k$, and NDCG@$k$ for $k \in \{1, 5, 10, 20\}$, averaged over each query split $\mathcal{Q} \in \{\mathcal{Q}_{\mathrm{eval}}, \mathcal{Q}_{\mathrm{vague}}\}$ and stratified by tier (G1/G2/G3). Catalog embeddings $\{f_{\theta}(\phi_{5}(t))\}_{t \in \mathcal{T}}$ are precomputed once per encoder $\theta$ under $\phi_{5}$ and reused across query splits; rewriter outputs are regenerated end-to-end for every reported configuration. Metric definitions appear in Appendix~\ref{app:metrics}; the full $k$-sweep results in Appendix~\ref{app:k_sweep}.

\section{Experiments \& Results}
\label{sec:experiments}

\subsection{Experimental Setup}
\label{sec:setup}

\paragraph{Benchmark and evaluation splits.}
All experiments use the ToolBench-derived catalog and query splits described in \S\ref{subsec:data-in-method}: a 10{,}000-tool subset $\mathcal{T}$ with 1{,}092 evaluation queries stratified across three tiers (G1/G2/G3).
Each query is evaluated on both the standard split $\mathcal{Q}_{\mathrm{eval}}$ — the original ToolBench queries — and the vague split $\mathcal{Q}_{\mathrm{vague}}$, which contains intent-preserving paraphrases that replace surface tokens with conversational alternatives (both splits share the same gold tool sets).

\paragraph{Baselines.}
We compare against seven reference points spanning the space of design choices.
\textbf{BM25} over the $\phi_{5}$-indexed catalog serves as a sparse lexical floor, requiring no training or LLM.
\textbf{BGE (vanilla)} and \textbf{text-embedding-3-large} are frozen dense encoders that embed raw queries directly.
\textbf{Query expansion (LLM + BGE)} and \textbf{HyDE (vanilla LLM + BGE)} both pair the same vanilla BGE encoder with the same vanilla Qwen3.5-4B generator, but differ in generation strategy: query expansion paraphrases the user query (anchor stays on the query side), while HyDE generates a hypothetical catalog-style tool description (anchor moves to the document side).
\textbf{BGE (trained S1a)} is the BGE encoder fine-tuned on (query, tool) pairs at the S1a warmup step described in \S\ref{sec:cotrain}.
\textbf{HyDE (vanilla LLM + trained BGE S1a)} pairs the trained encoder with HyDE generation without any rewriter training, testing whether the two components can be composed after independent optimisation.
All baselines use the $\phi_{5}$ catalog index for a fair comparison; all LLM-based baselines use Qwen3.5-4B~\citep{qwen3report} as the generator.

\paragraph{\cohyde{} inference.}
The trained rewriter produces $\tilde{d} = \mathrm{clean}(g_{\psi}(\rho_{\mathrm{HyDE}}(q)))$ (greedy, 150-token budget); the trained encoder retrieves top-$k$ tools from the $\phi_{5}$ index. Full hyperparameters are in Appendix~\ref{app:hp-summary}.

\paragraph{Metrics.}
NDCG@5 is the primary metric; Recall@5 is reported as a secondary check that gains reflect more correct tools being retrieved and not merely reranking an already-correct candidate set.
Both metrics are reported on $\mathcal{Q}_{\mathrm{eval}}$ and $\mathcal{Q}_{\mathrm{vague}}$, stratified by tier (G1\,/\,G2\,/\,G3), giving six (metric $\times$ split $\times$ tier) cells per configuration.

\subsection{\cohyde{} Comparison with Baselines}
\label{sec:results}

\begin{table*}[t]
\centering
\small
\setlength{\tabcolsep}{4pt}
\renewcommand{\arraystretch}{1.1}
\resizebox{\textwidth}{!}{%
\begin{tabular}{l cc cc cc cc cc cc}
\toprule
 & \multicolumn{6}{c}{Standard} & \multicolumn{6}{c}{Vague} \\
\cmidrule(lr){2-7} \cmidrule(lr){8-13}
 & \multicolumn{2}{c}{G1} & \multicolumn{2}{c}{G2} & \multicolumn{2}{c}{G3}
 & \multicolumn{2}{c}{G1} & \multicolumn{2}{c}{G2} & \multicolumn{2}{c}{G3} \\
\cmidrule(lr){2-3} \cmidrule(lr){4-5} \cmidrule(lr){6-7}
\cmidrule(lr){8-9} \cmidrule(lr){10-11} \cmidrule(lr){12-13}
Method & N@5 & R@5 & N@5 & R@5 & N@5 & R@5 & N@5 & R@5 & N@5 & R@5 & N@5 & R@5 \\
\midrule
BM25                                   & 49.0 & 52.7 & 31.0 & 31.7 & 17.3 & 17.5 & 17.8 & 19.6 & 10.3 & 11.6 &  3.4 &  3.0 \\
BGE (vanilla)                          & 56.5 & 62.4 & 31.9 & 33.2 & 18.5 & 20.7 & 33.2 & 37.3 & 15.8 & 16.7 &  8.3 & 12.9 \\
text-embedding-3-large                 & 61.6 & 66.4 & 36.3 & 36.4 & 26.7 & 29.2 & 34.8 & 40.5 & 18.1 & 19.8 & 19.8 & 19.8 \\
Query expansion$^\dagger$ (LLM + BGE)  & 51.4 & 56.7 & 27.0 & 30.0 & 29.2 & 30.9 & 32.6 & 38.2 & 11.3 & 12.5 &  6.2 & 10.9 \\
HyDE (vanilla LLM \& BGE)             & 58.2 & 63.8 & 34.9 & 37.7 & 35.1 & 36.4 & 37.4 & 43.4 & 17.7 & 19.9 & 17.4 & 21.1 \\
BGE (trained S1a)                      & 84.2 & 89.7 & 71.7 & 76.3 & 57.1 & 56.8 & 44.7 & 49.8 & 30.7 & 33.6 & 14.9 & 18.7 \\
HyDE (vanilla LLM + trained BGE S1a)  & 73.4 & 79.5 & 59.5 & 62.2 & 46.2 & 47.6 & 41.2 & 47.4 & 29.3 & 31.8 & 17.6 & 17.9 \\
\textbf{\cohyde{} ($r=3$)}             & \textbf{86.8} & \textbf{91.0} & \textbf{73.6} & \textbf{78.0} & \textbf{60.1} & \textbf{60.4} & \textbf{49.4} & \textbf{55.2} & \textbf{38.7} & \textbf{41.5} & \textbf{21.1} & \textbf{26.2} \\
\bottomrule
\end{tabular}}
\begin{minipage}{\textwidth}\small
$^\dagger$ Vanilla LLM paraphrases the user query into a retrieval-friendly form (query-side expansion); the rewritten query is encoded by vanilla BGE.
\end{minipage}
\caption{NDCG@5 (N@5) and Recall@5 (R@5) in \% on standard and vague query splits, stratified by tier. Bold = best per column.}
\label{tab:main}
\end{table*}

Table~\ref{tab:main} compares \cohyde{} against seven reference points; reading the rows top to bottom traces the logical sequence that motivates the co-training design.

\paragraph{Encoder-only fine-tuning is brittle on vague queries.}
The InfoNCE-trained encoder (BGE S1a) dominates every standard evaluation split by a wide margin, lifting G1 NDCG@5 from 56.5 to 84.2 over vanilla BGE.
On vague paraphrases of the same queries, however, it collapses: G1 vague falls $-39.5$pp from its own performance on standard counterpart, and G3 vague reaches $14.9$\% — barely above the vanilla baseline. The strong commercial encoder (text-embedding-3-large) follows the same pattern: competitive on standard, but no more robust on vague. The encoder has learned a similarity function calibrated to the surface vocabulary of well-formed queries; any deviation from that vocabulary exposes its brittleness.

\paragraph{Description generation bridges vocabulary gaps; query rewriting does not.}
Table~\ref{tab:main} includes both a query expansion baseline and a HyDE baseline, both using the same vanilla BGE encoder and the same vanilla Qwen3.5-4B generator.
On standard queries the two are comparable; the decisive difference is on vague cross-domain queries. Query rewriting, which keeps the inference-time anchor on the query side of the embedding space, reaches G3 vague NDCG@5 of only $6.2$\%---below the vanilla BGE baseline of $8.3$\%.
HyDE, which generates hypothetical catalog-style tool descriptions and moves the anchor to the document side, reaches $17.4$\% on the same split, a $+11.2$pp gap.
The pattern is consistent across all tiers: HyDE outperforms query rewriting on every vague cell, often by double-digit margins.
This establishes the generative direction that \cohyde{} adopts: producing a hypothetical tool description rather than reformulating the query.

\paragraph{Combining HyDE with a query-trained encoder makes things worse.}
A natural next step is to combine the gains of encoder fine-tuning with HyDE generation.
Table~\ref{tab:main} shows that this naive combination \emph{backfires}: ``HyDE (vanilla LLM + trained BGE S1a)'' drops $-10.8$pp on G1 standard NDCG@5 relative to the trained encoder used alone ($73.4$ vs $84.2$), and trails on every other split as well.
The trained encoder's similarity function was calibrated on raw user queries as anchors; at inference it receives hypothetical catalog descriptions whose embedding distribution is shifted away from that calibration manifold, distorting the nearest-neighbour search.
This is the direct motivation for co-training: the encoder and rewriter cannot be composed after independent training. Instead, they should evolve their representation spaces together.

\paragraph{\cohyde{} resolves all three failure modes simultaneously.}
\cohyde{} at $r{=}3$ improves over the strongest single-component baseline (BGE S1a) on every split.
Standard-query gains are modest (average $+2.5$pp), reflecting that co-training preserves the encoder's standard-query precision rather than trading it away.
Vague-query gains are substantially larger (average $+6.3$pp), closing the lexical brittleness that neither the trained encoder nor baseline HyDE could resolve on its own.
Crucially, the co-trained encoder also closes the representation-mismatch gap: trained exclusively on DPO-generated hypothetical descriptions with zero raw queries in its training data, it reaches G1 standard NDCG@5 of $86.8$\%, matching and slightly exceeding the BGE encoder trained on raw queries.
The jointly-trained space has been shaped so that raw query vectors at inference land in the same neighbourhood as their corresponding catalog descriptions, without ever having seen those queries during training.

\subsection{Ablations}
\label{sec:ablations}

\begin{table*}[t]
\centering
\small
\setlength{\tabcolsep}{4pt}
\renewcommand{\arraystretch}{1.1}
\resizebox{\textwidth}{!}{%
\begin{tabular}{l cc cc cc cc cc cc}
\toprule
 & \multicolumn{6}{c}{Standard} & \multicolumn{6}{c}{Vague} \\
\cmidrule(lr){2-7} \cmidrule(lr){8-13}
 & \multicolumn{2}{c}{G1} & \multicolumn{2}{c}{G2} & \multicolumn{2}{c}{G3}
 & \multicolumn{2}{c}{G1} & \multicolumn{2}{c}{G2} & \multicolumn{2}{c}{G3} \\
\cmidrule(lr){2-3} \cmidrule(lr){4-5} \cmidrule(lr){6-7}
\cmidrule(lr){8-9} \cmidrule(lr){10-11} \cmidrule(lr){12-13}
Variant & N@5 & R@5 & N@5 & R@5 & N@5 & R@5 & N@5 & R@5 & N@5 & R@5 & N@5 & R@5 \\
\midrule
\cohyde{} (full)                           & \textbf{86.8} & \textbf{91.0} & 73.6          & \textbf{78.0} & \textbf{60.1} & \textbf{60.4} & \textbf{49.4} & \textbf{55.2} & \textbf{38.7} & \textbf{41.5} & \textbf{21.1} & \textbf{26.2} \\
\midrule
\cohyde{} (w/o S1b rewriter warmup)        & 81.3          & 87.0          & 71.5          & 75.6          & 50.5          & 53.8          & 47.0          & 54.1          & 35.2          & 36.9          & 19.6          & 21.8          \\
\cohyde{} (trained LLM + vanilla encoder)  & 63.2          & 68.7          & 38.1          & 40.3          & 36.2          & 37.0          & 40.1          & 45.2          & 17.6          & 19.4          & 12.9          & 15.3          \\
\cohyde{} (vanilla LLM + trained encoder)  & 86.3          & 79.5          & \textbf{75.6} & 62.2          & 53.7          & 47.6          & 44.1          & 47.4          & 32.8          & 31.8          & 15.8          & 17.9          \\
\bottomrule
\end{tabular}}
\caption{Ablation study. Each row removes or replaces one component of \cohyde{}. Bold = best per column.}
\label{tab:ablation}
\end{table*}

\begin{figure}[t]
  \centering
  \includegraphics[width=\linewidth]{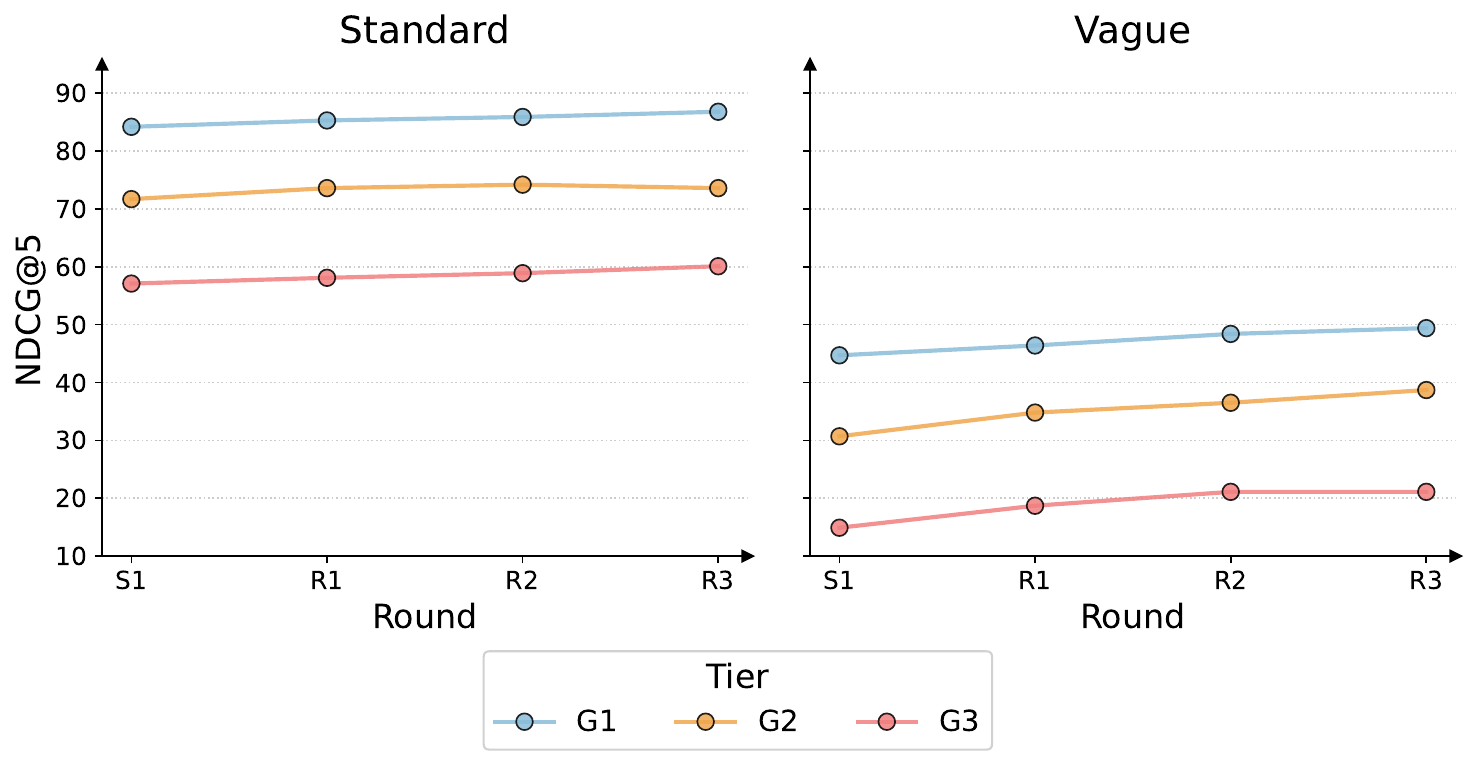}
  \caption{Per-round NDCG@5 trajectory on standard (left) and vague (right) query splits, stratified by tier. Performance improves monotonically on five of six splits from S1 through R3; the single exception is standard G2, which retreats by a marginal $0.6$\,pp between R2 and R3.}
  \label{fig:trajectory}
\end{figure}

We isolate four design choices in \cohyde{}: (i) the rewriter warmup stage S1b, which pre-trains the LLM on catalog surface forms before the co-training loop begins; (ii) the joint encoder update, asking whether the gains require a co-trained encoder or can be obtained by pairing the trained rewriter with a vanilla encoder; (iii) the symmetric question for the encoder side, asking whether the co-trained encoder retains its advantage when paired with a vanilla (untrained) rewriter; and (iv) the number of co-training rounds $r$, which measures convergence behaviour and whether additional rounds continue to improve retrieval quality.

Table~\ref{tab:ablation} reports results for each ablated variant across all six evaluation splits.

\paragraph{Rewriter warmup is critical for cross-domain retrieval.}
Removing the rewriter warmup drops standard G3 NDCG@5 by $9.6$pp ($60.1 \to 50.5$) and R@5 by $6.6$pp, while standard G1 and G2 fall by only $5.5$pp and $2.1$pp respectively.
Vague-query degradation is consistently smaller ($\leq 3.5$pp across all tiers).
The gradient of the drop, steepest on standard G3 and shallowest on vague splits, reflects what the warmup actually provides: the rewriter learns the catalog's vocabulary and surface forms \emph{before} the co-training loop begins. On near-domain standard G1 queries, the encoder can partially compensate for a cold rewriter; on cross-domain standard G3 tools, whose descriptions share few surface tokens with user queries, a warmup-free rewriter fails to generate catalog-aligned descriptions from the outset and the encoder's nearest-neighbour search degrades from round one.

\paragraph{The trained rewriter requires a jointly-trained encoder.}
Pairing the co-trained rewriter with a vanilla BGE encoder produces the largest degradation in Table~\ref{tab:ablation}. NDCG@5 collapses on standard splits by $23.6$pp, $35.5$pp, and $23.9$pp on G1, G2, and G3 respectively; vague splits decline by $9$--$21$pp. The vanilla encoder was trained on raw user queries, so its representation space is calibrated to natural-language vectors rather than to the catalog-style hypothetical descriptions the DPO-aligned rewriter generates. Feeding it rewriter outputs at inference therefore distorts, rather than improves, the similarity search. This result confirms that the rewriter's gains are not a free add-on to any encoder: they require an encoder whose representation space has been co-shaped to match the rewriter's output distribution.

\paragraph{The encoder is load-bearing for standard queries; the rewriter differentiates vague ones.}
The symmetric ablation, co-trained encoder with a vanilla rewriter, reveals the complementary side. On easy standard queries, the co-trained encoder is nearly self-sufficient: G1 standard NDCG@5 falls by only $0.5$pp ($86.3$ vs $86.8$), and G2 standard actually edges out the full model by $2.0$pp ($75.6$ vs $73.6$). The co-trained encoder has absorbed enough of the catalog distribution that zero-shot HyDE queries land acceptably close in its embedding space without a fine-tuned rewriter. The gap opens on harder settings: NDCG@5 on standard G3 falls by $6.4$pp ($60.1 \to 53.7$) and on vague splits by $5.3$--$5.9$pp uniformly across all tiers. These are precisely the conditions where the rewriter's DPO alignment matters---bridging a large lexical gap on cross-domain tools, or reasoning past underspecification on vague queries.

Together, ablations (ii) and (iii) confirm the asymmetry established in \S\ref{sec:results}: the encoder carries precision on near-vocabulary standard queries; the rewriter provides robustness on hard and vague ones; co-training is what enables both gains simultaneously.

\paragraph{Co-training performance evolution across rounds.}
Figure~\ref{fig:trajectory} traces NDCG@5 at each stage for all six evaluation splits. Performance is monotonically non-decreasing from S1 through R3 on five of six splits; the single exception is standard G2, which retreats by a marginal $0.6$pp between R2 and R3. Gains from R1 to R2 are consistently larger than those from R2 to R3 across all tiers and both evaluation split types. A diagnostic round R4 (seed~123) reveals that the system does not simply plateau: standard-split NDCG@5 continues to rise ($+5.0$\,pp on G3: $57.5 \to 62.5$), while all three vague splits decline ($-0.9$, $-1.8$, $-0.9$\,pp on G1/G2/G3; full table in Appendix~\ref{app:seeds}). R3 therefore marks the round at which vague-query performance peaks; reporting it reflects a robustness--precision trade-off rather than an arbitrary early stop. We emphasize that $R$ is selected purely on the dev-split NDCG@5 defined in \S\ref{subsec:data-in-method}, never on the 1{,}092-query test set used throughout this section; re-verifying round selection under this dev-only procedure confirms that $R{=}3$ remains the selected round and still improves over BGE~(trained S1a) on all six evaluation cells in Table~\ref{tab:main}.

\subsection{Multi-Seed Validation}
\label{sec:seeds}

The results in \S\ref{sec:results}--\S\ref{sec:ablations} are reported under a single training seed (seed~0). To check that the reported gains are not an artifact of one run, we repeat the full co-training pipeline (S1a through R3) under an independent seed (123), with the S1a/S1b warmup stages held fixed since only the co-training loop differs across seeds.

\begin{table}[t]
\centering
\small
\setlength{\tabcolsep}{3pt}
\renewcommand{\arraystretch}{1.05}
\begin{tabular}{ll ccc ccc}
\toprule
 & & \multicolumn{3}{c}{Standard NDCG@5} & \multicolumn{3}{c}{Vague NDCG@5} \\
Seed & Stage & G1 & G2 & G3 & G1 & G2 & G3 \\
\midrule
0   & S1 & 84.2 & 71.7 & 57.1 & 44.7 & 30.7 & 14.9 \\
    & R1 & 85.3 & 73.6 & 58.1 & 46.4 & 34.8 & 18.7 \\
    & R2 & 85.9 & 74.2 & 58.9 & 48.4 & 36.5 & 21.1 \\
    & R3 & 86.8 & 73.6 & 60.1 & 49.4 & 38.7 & 21.1 \\
\midrule
123 & R1 & 86.6 & 73.7 & 50.6 & 46.8 & 34.8 & 17.9 \\
    & R2 & 86.7 & 75.7 & 57.2 & 47.2 & 35.2 & 20.3 \\
    & R3 & 88.0 & 76.9 & 57.5 & 48.1 & 37.7 & 20.8 \\
\bottomrule
\end{tabular}
\caption{NDCG@5 per round under both training seeds. Seed~0 is the main paper's run; seed~123 is an independent rerun. S1 values are not re-run for seed~123 (the S1a/S1b warmup is identical across seeds; only the co-training loop differs).}
\label{tab:seeds}
\end{table}

Table~\ref{tab:seeds} reports per-round NDCG@5 under both seeds across all six evaluation splits. Final-round (R3) deltas between seeds fall within the paired-bootstrap CI half-widths for each tier ($\pm2$\,pp G1, $\pm3$\,pp G2, $\pm5$--$6$\,pp G3; Appendix~\ref{app:bootstrap}), and the qualitative pattern established in \S\ref{sec:results}---\cohyde{} improving over the strongest single-component baseline (BGE, trained S1a) on every split---holds under both seeds. Under seed~123, every evaluation cell is in fact monotonically non-decreasing across all three rounds, unlike seed~0's single non-monotonic cell (standard G2, \S\ref{sec:ablations}), suggesting that irregularity is training noise rather than a systematic property of the method. A further round (R4, seed~123) run purely as a convergence diagnostic is detailed in Appendix~\ref{app:seeds}: standard-split NDCG@5 continues to improve at R4 while all three vague splits decline, indicating that R3---not a training plateau---marks the point of peak vague-query robustness.

\subsection{Comparison with Closest Prior Methods}
\label{sec:prior_comparison}

\cohyde{} is most directly related to two lines of work that also use iterative feedback to improve tool or document retrieval. \citet{xu2024iterative} propose an iterative loop in which the LLM's \emph{downstream tool-usage success} is fed back to retrain the retriever; the retriever evolves across rounds but the query representation at inference is the raw user query and no rewriter component is trained. RaFe~\citep{mao2024rafe} trains a query rewriter with RL feedback from an external reranker in a general RAG setting; critically, the rewriter \emph{paraphrases the user query} into a more retrieval-friendly form — it does not generate catalog-style hypothetical descriptions — so the inference-time anchor remains on the query side of the embedding space, and the encoder remains frozen throughout.
Both methods train only one side of the retrieval pipeline and use a signal external to the encoder-rewriter pair rather than closing the loop directly through the retrieval objective.

\begin{figure}[t]
  \centering
  \includegraphics[width=\linewidth]{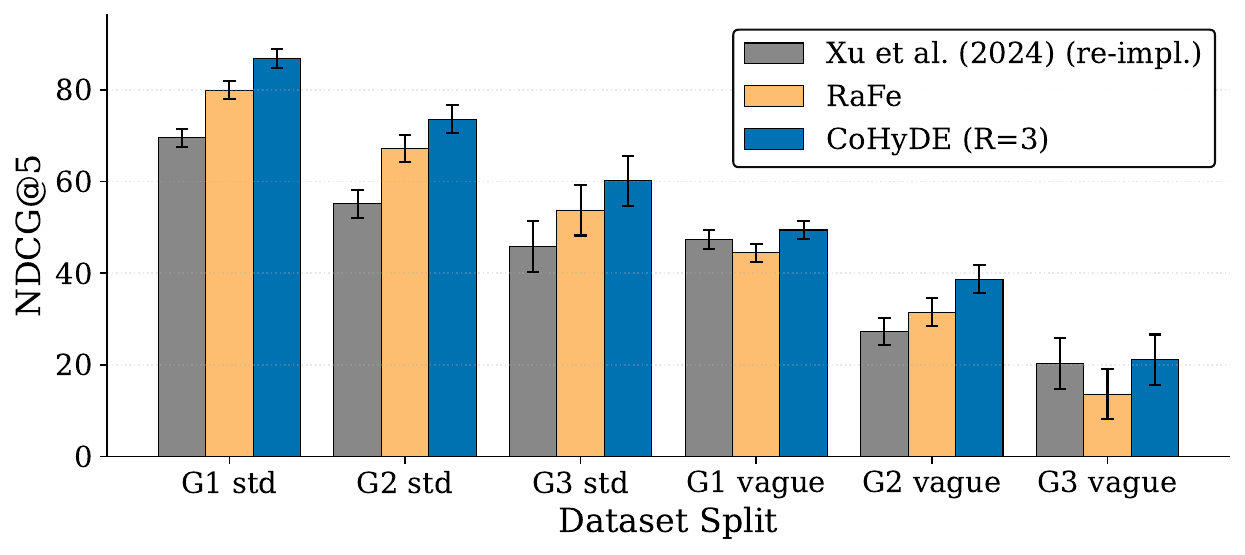}
  \caption{NDCG@5 comparison with the two closest prior methods across all six evaluation splits. Error bars show 95\% confidence intervals.}
  \label{fig:head_to_head}
\end{figure}

Figure~\ref{fig:head_to_head} reports NDCG@5 for all three methods across standard and vague splits. On standard queries \cohyde{} leads on all three tiers by a wide margin over \citet{xu2024iterative}: $+17.3$pp on G1, $+18.5$pp on G2, and $+14.3$pp on G3. RaFe~\citep{mao2024rafe} is a stronger standard-query competitor than \citet{xu2024iterative}, closing much of the gap, but still trails \cohyde{} by $6.9$pp on G1, $6.4$pp on G2, and $6.4$pp on G3. The vague splits separate the methods more sharply. \citet{xu2024iterative} holds up on G1 and G2 vague (within $\leq 2.1$pp of \cohyde{}), but falls behind on G3. RaFe~\citep{mao2024rafe} degrades most severely on G3 vague, dropping to $13.6$ NDCG@5 against \cohyde{}'s $21.1$ --- a $7.5$pp gap on the hardest cross-domain vague split, compared to RaFe's $6.4$pp deficit on the corresponding standard split. Confidence intervals for all reported differences follow the paired-bootstrap protocol (Appendix~\ref{app:bootstrap}); re-implementation details for \citet{xu2024iterative} are in Appendix~\ref{app:xu-repro}.

\cohyde{}'s joint co-training sustains this advantage: the encoder's retrieval metric directly supervises the rewriter, and the rewriter's outputs shape the encoder's representation space, closing the lexical gap that both prior methods leave open.

\section{Conclusion}
\label{sec:conclusion}

Contrastive encoder fine-tuning and HyDE-style description generation fail in complementary directions, and their naive composition makes things worse because their representation spaces have been calibrated to different input distributions.
We introduce \cohyde{}, an iterative co-training loop that resolves this by evolving the encoder and rewriter together: the encoder's NDCG@5 scores supervise the rewriter via DPO, and the rewriter's catalog-aligned outputs become the encoder's training anchors each round.
Three rounds improve over the strongest single-component baseline on every evaluation cell, with average gains of $+2.5$pp NDCG@5 on standard queries and $+6.3$pp on vague ones.
The asymmetric improvement reflects the mechanism directly: the jointly-trained encoder maps raw queries near catalog descriptions at inference without ever seeing those queries during training, confirming that the encoder and rewriter are better treated as a single co-evolving system.

\section*{Limitations}
\label{sec:limitations}

We validated the main results under a second independent seed (123); final-round NDCG@5 differences between seeds fall within the paired-bootstrap CI half-widths for each tier ($\pm 2$\,pp G1, $\pm 3$\,pp G2, $\pm 5$--$6$\,pp G3), and the qualitative pattern---\cohyde{} improving over the strongest single-component baseline on every split---holds under both seeds. Full cross-seed NDCG@5 trajectories are reported in \S\ref{sec:seeds}; the R4 convergence diagnostic is in Appendix~\ref{app:seeds}.
Experiments are conducted on a 10K-tool English subset of ToolBench, which is skewed toward consumer-facing RapidAPI REST endpoints; it remains to be seen whether the co-training gains transfer to enterprise catalogs, non-English queries, or function-call schemas that lack free-text descriptions.
The vague-query split $\mathcal{Q}_{\mathrm{vague}}$ is generated and validated by the LLMs used throughout the pipeline; though spot-checked by a human on 50 paraphrases (Appendix~\ref{app:vague}), systematic biases shared between the generator and judge may go undetected.
Finally, we benchmark against single-vector dense retrievers and BM25 but not against cross-encoder rerankers or sparse--dense hybrids; a comparison with such methods would require matched latency or FLOPs budgets, which we leave to future work.

\section*{Ethical Considerations}
\label{sec:ethics}

We conducted experiments within the provisions of the ACL Ethics Policy and relevant research-integrity guidelines. There are, to the best of our knowledge, no remaining ethical risks that have not been addressed.

\bibliography{custom}

\appendix
\section*{Appendix}


\section{Vague-Query Construction and Validation}
\label{app:vague}

The vague-query split $\mathcal{Q}_{\mathrm{vague}}$ is a held-out paraphrase of the official 1{,}092-query evaluation set, constructed to probe robustness under query-side distribution shift while preserving the gold tool set. Construction follows the protocol of \citet{chen2026trb} verbatim, with claude-4.5-opus substituted for their GPT-4o paraphraser.

\paragraph{Two-pass validation.} The split is validated in two passes.
\begin{enumerate}
\item \textbf{LLM self-check.} Every paraphrase in $\mathcal{Q}_{\mathrm{vague}}$ is re-presented to claude-4.5-opus in a separate session, together with the original query and the gold tool set, and scored on the three binary criteria of \citet{chen2026trb} --- (i) intent preservation, (ii) absence of leaked tool names / API verbs / domain keywords, (iii) plausibility as an end-user utterance. An example is retained only if all three criteria are satisfied. Substituting claude-4.5-opus for the GPT-4o validator used by \citet{chen2026trb} is the only deviation from their protocol.
\item \textbf{Human spot-check.} 50 paraphrases were sampled uniformly at random from the LLM-validated split and re-verified by human against the same three criteria. All 50 passed all three criteria, giving a 6\% rule-of-three upper bound on the true failure rate at 95\% confidence. The annotator was not blinded to the paraphraser identity; this is a transparency disclosure rather than a methodological strength. The annotator was not compensated separately, apart from their regular wages during the research; no external annotators were used.
\end{enumerate}

\paragraph{Ethics review.} The annotation involves no human subjects beyond the human conducting the spot-check and qualifies as exempt from formal ethics-board review under the relevant institutional guidelines. No consent procedure was required because the annotator is the data producer.

\paragraph{Annotator demographics.} The annotator is a full-time employee of the authoring organization and resides in the USA.


\section{Cleaning Operator}
\label{app:clean}

The deterministic cleaning operator $\mathrm{clean}(\cdot)$ applied to every rewriter output before encoding strips:
\begin{enumerate}
    \item Reasoning-trace blocks delimited by \texttt{<think>...</think>}.
    \item Unclosed reasoning traces (a leading \texttt{<think>} with no terminator), in which case the entire output is rejected and replaced with the original query.
    \item Conversational preambles matching \texttt{\textasciicircum(Sure|Okay|Of course|Here is|Here's)[\textasciicircum.]*$\backslash$.$\backslash$s+}.
    \item Trailing whitespace and repeated blank lines.
\end{enumerate}
The operator is implemented as a sequence of regular-expression substitutions and is applied identically at SFT-target construction, DPO-candidate scoring, and inference time.

\section{HyDE-Style Rewriter Prompt}
\label{app:hyde-prompt}

The HyDE-style prompt $\rho_{\mathrm{HyDE}}$ is used at the optional SFT stage (when included; see Appendix~\ref{app:sft-stage}), at S2 (generating $\mathcal{D}_{\mathrm{d}}^{(\psi_{r})}$), at S4 (sampling DPO candidates), and at every inference-time HyDE evaluation reported in \S\ref{sec:experiments}.

\paragraph{System message:}
\begin{quote}\small
You are an expert at understanding API tool pipelines. When given a user query, you describe the sequence of API calls needed to fulfill it. Each description should focus on what the tool does, what inputs it takes, and what data it returns. Write each tool's description as a single concise technical sentence.
\end{quote}

\paragraph{User message:}
\begin{quote}\small\ttfamily
User query: \{query\}\\[4pt]
Think about the full pipeline of API calls needed to answer the query. Describe each API tool in the pipeline in order, explaining what data it provides and how it feeds into the next step. Be concise and technical.
\end{quote}

\section{Query-Rewriting Prompt}
\label{app:rewrite-prompt}

The query-rewriting prompt $\rho_{\mathrm{rewrite}}$ is used \emph{only} in the prompt-style ablation reported in Appendix~\ref{app:ablations}; it is not used anywhere for \cohyde{}.

\paragraph{System message:}
\begin{quote}\small
You are a query enhancement expert. Given a user query and the relevant API tools, rewrite the query to be more specific and detailed. Include relevant tool names, API capabilities, and technical terms that would help a retrieval system find the right tools. Keep it as a natural user request, but a more specific version of what the user is asking for.
\end{quote}

\paragraph{User message:}
\begin{quote}\small\ttfamily
Original query: \{query\}\\[4pt]
Relevant tools: \{tool\_names\}\\[4pt]
Rewritten query:
\end{quote}


\section{Per-Stage Hyperparameter Summary}
\label{app:hp-summary}

Table~\ref{tab:hp-summary} consolidates every training and inference stage in the main pipeline with its load-bearing hyperparameters. Full per-stage detail (objective, optimiser, schedules, ablation context) is in Appendix~\ref{app:encoder-hp} (encoder; S1a, S3$_{r}$) and Appendix~\ref{app:rewriter-hp} (rewriter; S1b, S2, S4$_{r}$). The optional HyDE-style SFT bridging stage --- which is \emph{not} part of the main pipeline --- is documented separately in Appendix~\ref{app:sft-stage}. Software versions for every stage are in Appendix~\ref{app:software}.

\begin{table*}[t]
\centering
\tiny
\setlength{\tabcolsep}{4pt}
\renewcommand{\arraystretch}{1.1}
\begin{tabular}{l l l l l l l}
\toprule
Stage & Objective & Init from & LR / $\beta$ & Effective BS & Length / steps & Adapter \\
\midrule
S1a (encoder)        & InfoNCE, $\tau{=}0.05$ & BGE-large-en-v1.5 & $2{\times}10^{-5}$ & 256 & seq 256, 5 ep & full FT \\
S1b (rewriter)       & 5-format SFT (NLL)     & Qwen3.5-4B          & $2{\times}10^{-5}$ & 64  & seq 1024, 8 ep & LoRA $r{=}16$ + new tok embeds \\
S2 (gen)             & greedy decode, $T{=}0$  & $\psi_{1}$        & ---                & 1   & 150-tok budget & --- \\
S3$_{r}$ (encoder)   & InfoNCE, $\tau{=}0.05$  & $\theta_{r}$      & $2{\times}10^{-5}$ & 256 & seq 256, 5 ep  & full FT \\
S4$_{r}$ (cand. sample) & nucleus, $T{=}0.7$, $p{=}0.95$ & $\psi_{r}$ & ---           & $N{=}4$/query & 300-tok budget & --- \\
S4$_{r}$ (DPO)       & sigmoid DPO, $\beta{=}0.1$ & $\psi_{r}$ (ref) & $5{\times}10^{-6}$ & 8 & prompt 1024 / comp 300, 1 epoch & LoRA $r'{=}64$ \\
Inference            & greedy decode, $T{=}0$  & $\psi_{R}$        & ---                & 1   & 150-tok budget & --- \\
\bottomrule
\end{tabular}
\caption{Per-stage hyperparameters for the main pipeline. ``LR'' is the optimiser learning rate (AdamW, weight decay $10^{-2}$, bf16 throughout); for S4 ``$\beta$'' is the DPO regularisation coefficient. ``Effective BS'' is per-device batch size $\times$ gradient accumulation. All training runs on a single H200 GPU; full per-stage detail in Appendices~\ref{app:encoder-hp},~\ref{app:rewriter-hp}.}
\label{tab:hp-summary}
\end{table*}

\section{Encoder Training Hyperparameters}
\label{app:encoder-hp}

The encoder is trained at two distinct points in the pipeline: once at S1a (warmup on real query--tool pairs) and once per round at S3$_{r}$ (retrain on rewriter-generated descriptions). Both stages use the same InfoNCE objective; they differ only in the anchor source and in whether they continue from the previous checkpoint or restart from $\theta_{0}$.

\paragraph{InfoNCE loss.} Let $\mathcal{B} = \{(a_{i}, p_{i})\}_{i=1}^{B}$ be a mini-batch of (anchor, positive) pairs, and write $S^{\theta}_{ij} = \langle f_{\theta}(a_{i}), f_{\theta}(p_{j}) \rangle / \tau$. The symmetric InfoNCE loss \citep{oord2018cpc} is
\begin{multline}
\mathcal{L}_{\mathrm{NCE}}(\theta; \mathcal{B}) = -\frac{1}{2B}\sum_{i=1}^{B}\!\Biggl[ \log\frac{\exp S^{\theta}_{ii}}{\sum_{j=1}^{B}\exp S^{\theta}_{ij}} \\
+ \log\frac{\exp S^{\theta}_{ii}}{\sum_{j=1}^{B}\exp S^{\theta}_{ji}} \Biggr],
\end{multline}
with temperature $\tau = 0.05$. Negatives are in-batch (no hard-negative mining).

\paragraph{S1a: Encoder warmup.} Anchors are real queries $q$, positives are $\phi_{5}(t)$ for the gold tool. Initialised from BGE-large-en-v1.5. AdamW with learning rate $\eta_{\theta} = 2 \times 10^{-5}$, weight decay $10^{-2}$, cosine schedule with 5\% warmup, batch size $B = 256$, max sequence length 256 tokens, 5 epochs over the 104{,}224 (G1+G2+G3) training pairs of $\mathcal{D}_{\mathrm{train}}$. Validation NDCG@5 (mean over G1/G2/G3 dev splits) is computed every 200 steps and the checkpoint maximising it is retained --- this is step 3600 in our run. All training is on a single H200 GPU with native bf16 mixed precision; no gradient accumulation. CLS-token pooling, L$_2$-normalised before scoring. No dropout beyond BGE's defaults.

\paragraph{S3$_{r}$: Per-round encoder retrain.} Anchors are the rewriter outputs $\tilde{d} = g_{\psi_{r}}(\rho_{\mathrm{HyDE}}(q))$ from the regenerated bootstrap set $\mathcal{D}^{(\psi_{r})}_{\mathrm{d}}$; positives are $\phi(t)$ for $\phi \sim \mathrm{Unif}(\Phi)$. The encoder is initialised from $\theta_{r}$ (i.e.\ continued from the previous round's checkpoint, not from $\theta_{0}$). All other hyperparameters --- optimiser, learning rate $2 \times 10^{-5}$, weight decay, cosine schedule with 5\% warmup, batch size 256, max sequence length 256, bf16, validation cadence, single-GPU --- are identical to S1a. The retrain runs for the same 5-epoch budget over $\mathcal{D}^{(\psi_{r})}_{\mathrm{d}}$, with the best validation-NDCG@5 checkpoint retained (around step 3400--4000 across rounds). \emph{No real $(q,t)$ pair is used at S3$_{r}$;} the encoder is trained purely on (rewritten-description, tool) pairs and tested on real queries at inference time. An ablation that mixes real $q$-anchored and $\tilde{d}$-anchored pairs in the same retrain is reported in Appendix~\ref{app:ablations} (combined-pair encoder retrain).

\section{Rewriter Training and Inference Hyperparameters}
\label{app:rewriter-hp}

The rewriter is trained at S4$_{r}$ (DPO alignment, run once per round), and is sampled from at S2 (bootstrap data generation), at S4$_{r}$ (DPO candidate sampling), and at inference time. Each of these uses different decoding settings, listed below.

\paragraph{S1b: 5-format tool-rendering SFT.}
The rewriter $\psi_{0} =$ Qwen3.5-4B is fine-tuned on the catalog $\mathcal{T}$ rendered under all five formats $\phi_{1}, \ldots, \phi_{5}$ (defined in \S\ref{subsec:data-in-method}). Each tool is presented as a next-token prediction target under each of the five rendering conventions, sampled with equal weight per mini-batch.
LoRA with rank $r = 16$, $\alpha_{\mathrm{LoRA}} = 32$, dropout 0.05, applied to attention $q,k,v,o$ projections. AdamW with learning rate $\eta_{\psi}^{\mathrm{S1b}} = 2 \times 10^{-5}$, linear schedule with 3\% warmup, per-device batch size 2, gradient accumulation 32 (effective batch size 64), max sequence length 1024, 8 epochs over the mixture ($\approx$50K examples per epoch), bf16 mixed precision, gradient checkpointing on (non-reentrant), single H200 GPU. Validation hit@5 on the G1/G2/G3 retrieval dev splits is computed every 100 steps and the best checkpoint is retained.

\paragraph{S2: Bootstrap description generation.} Using $\psi_{1}$ (the S1b checkpoint), we generate the first round of (description, tool) training data $\mathcal{D}^{(\psi_{1})}_{\mathrm{d}}$ over all queries $q \in \mathcal{D}_{\mathrm{train}}$. Sampling is greedy ($T = 0$, top-$p = 1$, top-$k = 1$, no repetition penalty) with a 150-token completion budget, served via vLLM. We use a single completion per query. Outputs pass through $\mathrm{clean}(\cdot)$ (Appendix~\ref{app:clean}) before being used as encoder anchors. The same generation protocol is re-run at the start of every subsequent round $r$ to produce $\mathcal{D}^{(\psi_{r})}_{\mathrm{d}}$ from the current rewriter $\psi_{r}$.

\paragraph{S4$_{r}$: DPO candidate sampling.} For each query $q \in \mathcal{D}_{\mathrm{train}}$ we sample $N = 4$ candidate descriptions from $\psi_{r}$ at temperature $T = 0.7$, top-$p = 0.95$, top-$k = 50$, with a 300-token completion budget, served via vLLM. Each candidate $\tilde{d}^{(j)}$ is encoded by the freshly-retrained encoder $\theta_{r+1}$ (from S3$_{r}$); candidates are scored by their NDCG@5 against the gold tool set $T^{*}_{q}$ under $\theta_{r+1}$. The chosen / rejected pair $(\tilde{d}^{+}_{q}, \tilde{d}^{-}_{q})$ is the (argmax, argmin) of the four scores. Queries whose four candidates yield identical NDCG@5 are dropped from the DPO set.

\paragraph{S4$_{r}$: DPO training.} We use TRL's \texttt{DPOTrainer} with the sigmoid loss formulation \citep{rafailov2023dpo}:
\begin{equation*}
\mathcal{L}_{\mathrm{DPO}}(\psi; \psi_{r}) = -\log\sigma\!\Bigl(\beta\bigl[\Delta_{\psi}(\tilde{d}^{+}, q) - \Delta_{\psi}(\tilde{d}^{-}, q)\bigr]\Bigr),
\end{equation*}
with $\Delta_{\psi}(\tilde{d}, q) = \log\frac{p_{\psi}(\tilde{d}|\rho_{\mathrm{HyDE}}(q))}{p_{\psi_{r}}(\tilde{d}|\rho_{\mathrm{HyDE}}(q))}$ and $\beta = 0.1$. LoRA with rank $r' = 64$, $\alpha_{\mathrm{LoRA}} = 128$, dropout 0.05, applied to attention $q,k,v,o$ projections only; embeddings and the language-model head are \emph{not} tuned at S4 (the new tool tokens are already learned at S1b and held fixed thereafter). AdamW with learning rate $\eta_{\psi}^{\mathrm{S4}} = 5 \times 10^{-6}$, cosine schedule with 3\% warmup, per-device batch size 2, gradient accumulation 4 (effective batch size 8), max prompt length 1024 tokens, max completion length 300 tokens, 1 epoch over the DPO pair set ($\approx$4{,}371 optimiser steps), bf16 mixed precision, gradient checkpointing on, single H200 GPU. The reference policy $\psi_{r}$ is the previous round's rewriter; at $r=1$ this is $\psi_{1}$ from S1b. The trained adapter is merged back into the base weights at the end of the round before $\psi_{r+1}$ is used at S2 of round $r+1$.

\paragraph{Inference-time decoding.} At evaluation time the rewriter is sampled greedily ($T = 0$) with a 150-token completion budget, single completion per query, served via vLLM. The deterministic cleaning operator $\mathrm{clean}(\cdot)$ (Appendix~\ref{app:clean}) is applied before the description is passed to the encoder. The same decoding protocol is used for both standard and vague evaluation passes.

\paragraph{Reference policy and adapter merging.} At each round $r$, the DPO reference $\psi_{r}$ is loaded from the merged checkpoint of round $r{-}1$ (or from $\psi_{1}$ at $r=1$). After DPO training, the LoRA adapter is merged into the base weights to produce $\psi_{r+1}$, which serves both as the next round's S2 generator and as the next round's DPO reference.

\paragraph{Implementation.} PyTorch with HuggingFace Transformers; encoder training uses an in-house InfoNCE script; rewriter SFT and DPO use TRL's \texttt{SFTTrainer} and \texttt{DPOTrainer} with PEFT for LoRA adapters. vLLM serves the rewriter at S2, S4 candidate sampling, and inference. Exact software versions are listed in Appendix~\ref{app:software}.

\section{Optional SFT Stage (HyDE-Style Bridging)}
\label{app:sft-stage}

This appendix describes an \emph{optional} HyDE-style SFT pass that we ran in early experiments but \emph{do not} use in the main pipeline reported in \S\ref{sec:experiments}. It is a separate stage from the 5-format tool-rendering SFT (S1b) used in the \cohyde{} pipeline.

In early experiments we inserted a brief LoRA-SFT pass between S1 and S2$_0$ to align the rewriter's output style with the catalog rendering. The motivation was that the base Qwen3.5-4B rewriter under $\rho_{\mathrm{HyDE}}$ produced free-form text whose length and style differed visibly from the 5-format catalog rendering --- in particular, outputs were often substantially longer than any single $\phi_i$. A short SFT pass on cleaned descriptions taught $\psi$ the catalog-style output vocabulary and stop tokens, narrowing this style gap.

Once we adopted the 5-format encoder warmup (S1) and the 5-format rewriter warmup S1b (Appendix~\ref{app:rewriter-hp}), the picture changed. By training the encoder under $\phi \sim \mathrm{Unif}(\Phi)$ --- with $\phi_5$ in particular being the long, multi-sentence rendering closest in length and style to the rewriter's output --- the encoder learns a representation that is approximately invariant across the style gap this SFT pass was originally designed to close, and S1b then teaches the rewriter the catalog vocabulary directly. In this configuration, S2 can be initialised from $\psi_{1}$ (the S1b checkpoint) without an intervening HyDE-style SFT pass, and the iterative loop proceeds as in Algorithm~\ref{alg:cotrain}.

\paragraph{Pipeline with optional SFT.} When included, the optional SFT pass produces an alternative $\psi_{1}$ as follows. Sampling descriptions from $\psi_{0}$ under $\rho_{\mathrm{HyDE}}$ for queries in $\mathcal{D}_{\mathrm{train}}$ and applying $\mathrm{clean}(\cdot)$ with a length filter ($|\tilde{d}| > 30$ characters) yields a cleaned set $\mathcal{S}_{\mathrm{SFT}}$ of $\approx$2{,}754 pairs. We then run a \emph{short} LoRA SFT pass --- explicitly \emph{not} trained to convergence:
\begin{equation}
\psi_{1}^{\mathrm{(opt)}} = \arg\min_{\psi}\, -\!\!\sum_{(q,\tilde{d})\in\mathcal{S}_{\mathrm{SFT}}} \log p_{\psi}\!\bigl(\tilde{d}\,\big|\,\rho_{\mathrm{HyDE}}(q)\bigr).
\end{equation}
The iterative loop then runs from $\psi_{1}^{\mathrm{(opt)}}$ instead of $\psi_{1}$ from S1b.

\paragraph{Hyperparameters.} LoRA with rank $r = 16$, $\alpha_{\mathrm{LoRA}} = 32$, dropout 0.05, applied to all attention projection matrices ($q,k,v,o$); embeddings and the language-model head are not tuned. AdamW with $\eta_{\psi} = 2 \times 10^{-4}$, linear schedule with 20-step warmup, effective batch size 32 (per-device 4 $\times$ gradient accumulation 8), max sequence length 512, 100 optimisation steps total ($\approx$3{,}200 examples seen, less than 2 epochs over $\mathcal{S}_{\mathrm{SFT}}$). bf16 mixed precision, single H200 GPU, gradient checkpointing on. Longer schedules (1{,}000 / 5{,}000 steps) degraded downstream DPO performance by reducing the diversity of candidates available to the S4 sampler; this ablation is reported in Appendix~\ref{app:ablations}.

\paragraph{Ablation.} \S\ref{sec:experiments} reports retrieval numbers for the main pipeline (without the optional SFT pass, i.e.\ S2 initialised from S1b). The variant with the optional SFT pass is reported in Appendix~\ref{app:ablations} and does not improve over the main pipeline.


\section{Evaluation Metrics}
\label{app:metrics}

For a query $q$ with gold tool set $T^{*}_{q}$ and retrieved ranking $\hat{T}_{k}(q) = (\hat{t}_{1}, \ldots, \hat{t}_{k})$:
\begin{align}
\mathrm{hit@}k(q) &= \mathbb{1}\!\bigl[\hat{T}_{k}(q) \cap T^{*}_{q} \neq \varnothing\bigr], \\
\mathrm{recall@}k(q) &= \frac{|\hat{T}_{k}(q) \cap T^{*}_{q}|}{|T^{*}_{q}|}, \\
\mathrm{NDCG@}k(q) &= \frac{\sum_{j=1}^{k} \frac{\mathbb{1}[\hat{t}_{j} \in T^{*}_{q}]}{\log_{2}(j+1)}}{\sum_{j=1}^{\min(k, |T^{*}_{q}|)} \frac{1}{\log_{2}(j+1)}}.
\end{align}
Each metric is averaged over queries in the relevant tier. Definitions match the standard \texttt{ir\_measures} implementations.

\section{Round-3 $k$-Sweep}
\label{app:k_sweep}

Table~\ref{tab:k_sweep} reports hit@$k$, recall@$k$, and NDCG@$k$ for the converged round-3 co-trained system at $k \in \{1, 5, 10, 20\}$, on both standard and vague query splits, stratified by tier. Numbers are sourced from the same evaluation run that supplies the round-3 NDCG@5. NDCG@1 equals hit@1 by construction. Recall@1 is reported in full but, as noted in \S\ref{sec:eval}, is bounded above by $1/|T^{*}_{q}|$ and is therefore lower than the other metrics for every multi-tool query.

\begin{table}[t]
\centering
\small
\setlength{\tabcolsep}{4pt}
\renewcommand{\arraystretch}{1.05}
\begin{tabular}{lccc|ccc}
\toprule
 & \multicolumn{3}{c|}{Standard} & \multicolumn{3}{c}{Vague} \\
Metric & G1 & G2 & G3 & G1 & G2 & G3 \\
\midrule
hit@1     & 83.1 & 75.2 & 69.0 & 44.7 & 39.4 & 18.0 \\
hit@5     & 95.8 & 90.7 & 92.0 & 66.1 & 65.2 & 58.0 \\
hit@10    & 97.3 & 92.0 & 96.0 & 75.7 & 73.9 & 74.0 \\
hit@20    & 98.0 & 93.0 & 98.0 & 82.1 & 80.0 & 86.0 \\
\midrule
recall@1  & 40.4 & 32.9 & 25.2 & 21.8 & 16.9 &  6.8 \\
recall@5  & 91.0 & 78.0 & 60.4 & 55.2 & 41.5 & 26.2 \\
recall@10 & 95.2 & 86.6 & 75.4 & 67.2 & 54.3 & 42.2 \\
recall@20 & 96.7 & 89.6 & 90.5 & 76.1 & 65.0 & 57.8 \\
\midrule
NDCG@1    & 83.1 & 75.2 & 69.0 & 44.7 & 39.4 & 18.0 \\
NDCG@5    & 86.8 & 73.6 & 60.1 & 49.4 & 38.7 & 21.1 \\
NDCG@10   & 87.8 & 78.3 & 66.8 & 54.4 & 44.0 & 28.0 \\
NDCG@20   & 88.3 & 78.4 & 72.4 & 57.2 & 47.5 & 33.6 \\
\bottomrule
\end{tabular}
\caption{Full $k$-sweep for the round-3 co-trained system. NDCG@1 = hit@1 by construction. Recall@1 is capped at $1/|T^{*}_{q}|$ for multi-tool queries.}
\label{tab:k_sweep}
\end{table}

\section{Bootstrap CI Protocol}
\label{app:bootstrap}

The paired-bootstrap 95\% confidence intervals reported in \S\ref{sec:experiments} and Appendix~\ref{app:seeds} are computed as follows. For each tier $G \in \{G_{1}, G_{2}, G_{3}\}$ and split $\in \{\mathrm{standard}, \mathrm{vague}\}$, let $\{x_{q}\}_{q \in \mathcal{Q}^{(G)}}$ and $\{y_{q}\}_{q \in \mathcal{Q}^{(G)}}$ be the per-query NDCG@5 scores under the two systems being compared (e.g.\ Round 3 and Xu re-implementation), each system having produced its own ranking for the same set of queries from the same evaluation run. We resample query indices with replacement, $B = 10{,}000$ times, with a fixed random seed. For each resample $b$, we compute $\bar{x}^{(b)} = \mathrm{mean}_{q \in S_{b}} x_{q}$ and $\bar{y}^{(b)} = \mathrm{mean}_{q \in S_{b}} y_{q}$, and the paired difference $\delta^{(b)} = \bar{x}^{(b)} - \bar{y}^{(b)}$. The 95\% CI of the difference is the (2.5\%, 97.5\%) percentile interval of $\{\delta^{(b)}\}_{b=1}^{B}$; we report this as $[\mathrm{lo}, \mathrm{hi}]$. The same protocol with a single-system $x$-only resample yields the per-method CI half-widths quoted in Appendix~\ref{app:seeds} ($\pm 2$pp / $\pm 3$pp / $\pm 5{-}6$pp on G1/G2/G3, dominated by tier size $|\mathcal{Q}^{(G)}|$).


\section{Design-Choice Ablations: Details}
\label{app:ablations}

This appendix gives the per-variant numbers and discussion behind the summary in \S\ref{sec:ablations}. None of these variants are part of the main pipeline.

\paragraph{Single-format encoder training} ($\phi \equiv \phi_{i}$ for a fixed $i$). Training under any single rendering matched the 5-format encoder on its matched evaluation rendering but underperformed on the others. Training under $\phi_{5}$ alone --- the rendering closest in length to the rewriter's output --- still produced an encoder less robust to rewriter outputs of varying length than the 5-format encoder. We interpret this as evidence that the format mixture is doing more than augmenting on the longest format: by forcing the encoder to assign similar embeddings to the same tool across five different surface forms, it learns a length- and style-invariant representation that the description-only S2 retrains can then build on.

\paragraph{Combined-pair encoder retrain.} Replacing the description-only S3$_{r}$ objective with the mixed batch $\mathcal{D}^{(\psi_{r});\alpha=0.5}_{\mathrm{q+d}}$ produced no improvement over description-only and slightly degraded vague-query performance. The mechanism we attribute this to is that mixing $q$-anchored pairs back into S3 partially pulls the encoder toward the on-distribution $q$-anchored fixed point established at S1 --- the very fixed point whose vague-query failure we are trying to escape. The description-only objective is, in this view, doing distribution shift on purpose.

\paragraph{Query-rewrite prompt $\rho_{\mathrm{rewrite}}$.} Substituting the catalog-style $\rho_{\mathrm{HyDE}}$ with the user-style $\rho_{\mathrm{rewrite}}$ at any stage of the loop (sampling SFT targets, generating S3$_{r}$ pairs, or sampling DPO candidates) lost on every standard metric, with the largest gap on cross-domain G3. The prompt is given the relevant tool names as in-context anchors, so the comparison is not a strawman: with that anchoring, $\rho_{\mathrm{rewrite}}$ produces specific, plausible user queries --- they simply do not match the style of the contrastive pair the encoder sees during S3 retraining.

\paragraph{Longer SFT schedules.} Extending the optional SFT pass (Appendix~\ref{app:sft-stage}) from 100 steps to 1{,}000 / 5{,}000 steps closed the SFT train loss but produced lower-diversity candidates at S4 and a smaller DPO margin, ultimately reducing the closed-loop gain. The DPO update relies on temperature-0.7 sampling spreading mass across distinguishable candidates; over-fitted rewriters concentrate that mass and degrade the preference signal.

\paragraph{HyDE-concat.} Concatenating $q$ with $\tilde{d}$ before encoding, in the spirit of Query2doc \citep{wang2023query2doc}, helped slightly on G1 standard but hurt vague queries --- where the original $q$'s lexical surface is precisely the surface we are trying to escape.

\section{Xu et al.\ 2024 Re-implementation}
\label{app:xu-repro}

This appendix documents the hyperparameters and prompts used for the head-to-head against \citet{xu2024iterative} reported in \S\ref{sec:prior_comparison}.

\paragraph{Encoder.} \citet{xu2024iterative}'s pipeline trains the encoder once contrastively and never updates it again at inference time. We instantiate this once-trained encoder with our S1a InfoNCE checkpoint (BGE-large-en-v1.5 fine-tuned with InfoNCE on real $(q, \phi_{5}(t))$ pairs; full hyperparameters in Appendix~\ref{app:encoder-hp}). This is a strictly stronger starting point than the Sentence-BERT base used in the original paper, and is therefore a charitable substitution: any gap our co-training closes against this baseline cannot be attributed to a weaker re-implemented encoder.

\paragraph{LLM refiner.} Qwen3.5-4B served via vLLM, identical model and serving setup as the main paper's rewriter (we deliberately use the same LLM as our rewriter to remove model-capacity confounds; \citet{xu2024iterative} use GPT-3.5). Greedy decoding ($T = 0$, top-$p = 1$, top-$k = 1$), max 400 generated tokens per stage. The same temperature is used at all three prompted stages within a round.

\paragraph{Iteration schedule.} $T = 3$ refinement rounds (matching the paper's reported best). Within each round, the LLM sees the current top-$K$ retrieved tools with $K = 10$ and runs the three-stage Comprehension / Assessment / Refinement prompts; the refined instruction (or N/A early-stop) becomes the next round's retrieval input. Final ranking is the last round's. Top-50 are saved for evaluation at $k \in \{1, 5, 10, 20\}$.

\paragraph{Three-stage prompts.} The paper does not provide verbatim text. Our reimplementation uses prompts that match the three-stage description in their \S3:
\begin{itemize}
\item \textbf{P\_Comprehension} (system $+$ user): summarise user goals and the functionalities of the top-$K$ retrieved tools, one short sentence per goal and per tool.
\item \textbf{P\_Assessment} (system $+$ user): given the comprehension and the retrieved set, decide which goals are SOLVED vs UNSOLVED and whether the ranking matches importance; output \texttt{Solved:} and \texttt{Unsolved:} sections.
\item \textbf{P\_Refinement} (system $+$ user): given the assessment, output either \texttt{N/A} (if all goals solved and ranking matches) or a refined one-paragraph instruction enriched with the missing intent.
\end{itemize}
This should be read as a faithful reimplementation of the pipeline structure rather than an exact reproduction of Xu's prompts.

\paragraph{Caveat.} Beyond the prompt approximation, our reimplementation differs from the original paper in two respects: (i) a stronger encoder (S1a InfoNCE BGE-large vs Sentence-BERT base), and (ii) a different LLM (Qwen3.5-4B vs GPT-3.5). Both substitutions advantage Xu's method on this benchmark, making the head-to-head charitable to it.


\section{Compute Budget and Infrastructure}
\label{app:compute}

\paragraph{Hardware.} All experiments were run on a single node with 8 H200 GPUs. The encoder training, rewriter SFT, rewriter DPO, and HyDE inference passes each fit on a single GPU; multi-GPU parallelism was used only opportunistically and not required for any reported result.

\paragraph{Per-stage wall-clock cost (approximate, single H200).}
\begin{itemize}
\item S1a (encoder InfoNCE warmup, 5 epochs, batch 256): $\sim$3 hours.
\item S1b (rewriter 5-format tool-memorisation SFT, $\approx$50K examples): $\sim$2 hours.
\item Per-round S2 (description regeneration over $\mathcal{D}_{\mathrm{train}}$ at $T=0$, 150-token budget, via vLLM): $\sim$2 hours.
\item Per-round S3 encoder retrain: $\sim$1.5 hours.
\item Per-round S4 DPO data generation ($N=4$ candidates per query at $T=0.7$, scored by the current encoder): $\sim$6 hours.
\item Per-round S4 DPO training ($\approx$4{,}371 steps, LoRA $r=64$): $\sim$4 hours.
\item Per-configuration vague-split evaluation (HyDE generation over 1{,}092 queries via vLLM): $\sim$2.5 hours per pass.
\end{itemize}

\paragraph{Inference latency (single H200, greedy decode, 150-token budget).}
Table~\ref{tab:latency} reports measured per-query wall-clock time for the full \cohyde{} inference pipeline. LLM generation accounts for $>97\%$ of total latency; encoding and ANN lookup are negligible. Any HyDE-style baseline in Table~\ref{tab:main} incurs the same inference-time graph (one generation call followed by one encoder forward pass), so this overhead is a property of HyDE-style inference in general. The latency cost over encoder-only retrieval (BGE S1a: $\approx$12\,ms/query, no generation) is shared by all generation-based methods.

\begin{table}[h]
\centering
\small
\begin{tabular}{lrrr}
\toprule
Component & Mean (ms) & p50 (ms) & p90 (ms) \\
\midrule
Rewriter generation & 534.5 & 524.5 & 525.5 \\
Encoder forward pass &  11.4 &  11.4 &  12.7 \\
ANN lookup (10K index) &   0.5 &   0.5 &   0.5 \\
\midrule
Total & 546.4 & 536.4 & 538.7 \\
\bottomrule
\end{tabular}
\caption{Per-query inference latency of \cohyde{} on a single H200.}
\label{tab:latency}
\end{table}

\paragraph{Total budget.} Three rounds of co-training plus all baselines, ablations, and rejected design-choice variants (Appendix~\ref{app:ablations}) totalled approximately 400--500 GPU-hours on H200-class hardware. Reproducing only the main result (S1a + S1b + three rounds + a single end-to-end vague evaluation) would take roughly 50 GPU-hours.

\section{Software Versions}
\label{app:software}

Encoder training uses an in-house InfoNCE script built on PyTorch 2.4 and HuggingFace Transformers 4.46. Rewriter SFT and DPO use TRL 0.11 (\texttt{SFTTrainer}, \texttt{DPOTrainer}) with PEFT 0.13 for LoRA adapters. Rewriter inference uses vLLM 0.6. Mixed-precision training uses native PyTorch bf16. Evaluation uses our own retrieval scoring code; metric definitions match the standard \texttt{ir\_measures} implementations and are given in closed form in Appendix~\ref{app:metrics}.

\section{R4 Convergence Diagnostic and Bootstrap CI Half-Widths}
\label{app:seeds}

The multi-seed validation table (per-round NDCG@5 under seed~0 and an independent seed~123 rerun) is reported in the main text as Table~\ref{tab:seeds} (\S\ref{sec:seeds}). This appendix reports the round-4 convergence diagnostic that follows from it, together with the bootstrap CI half-widths used to interpret the seed comparison.

\paragraph{R4 convergence diagnostic (seed~123).}
Table~\ref{tab:r4} reports an additional round beyond the main paper's R3. Standard-split performance continues to improve at R4, with a notable $+5.0$\,pp gain on G3 ($57.5 \to 62.5$~NDCG@5). All three vague splits decline from R3 to R4 ($-0.9$, $-1.8$, $-0.9$\,pp on G1/G2/G3 respectively). This indicates that R3 sits at the point where further co-training trades vague-query robustness for standard-query gains; reporting R3 is a principled choice rather than an arbitrary early stop.

\begin{table}[t]
\centering
\small
\setlength{\tabcolsep}{4pt}
\renewcommand{\arraystretch}{1.05}
\begin{tabular}{l cc cc cc}
\toprule
 & \multicolumn{2}{c}{G1} & \multicolumn{2}{c}{G2} & \multicolumn{2}{c}{G3} \\
Round & N@5 & R@5 & N@5 & R@5 & N@5 & R@5 \\
\midrule
\multicolumn{7}{l}{\textit{Standard queries}} \\
R3 & 88.0 & 92.5 & 76.9 & 79.9 & 57.5 & 61.0 \\
R4 & 88.1 & 92.1 & 77.6 & 81.4 & 62.5 & 65.2 \\
\midrule
\multicolumn{7}{l}{\textit{Vague queries}} \\
R3 & 48.1 & 54.3 & 37.7 & 39.6 & 20.8 & 24.5 \\
R4 & 47.2 & 52.9 & 35.9 & 38.8 & 19.9 & 21.3 \\
\bottomrule
\end{tabular}
\caption{R3 vs.\ R4 under seed~123. Standard splits improve; all vague splits decline, establishing R3 as the vague-query optimum.}
\label{tab:r4}
\end{table}

\paragraph{Bootstrap CI half-widths (seed~0).}
We computed paired-bootstrap 95\% CIs ($B{=}10{,}000$) of NDCG@5 over the 593\,/\,399\,/\,100 evaluation queries in G1/G2/G3 (protocol in Appendix~\ref{app:bootstrap}). The half-width of these CIs---which captures sampling uncertainty over the evaluation set, \emph{not} training-seed variance---is approximately $\pm 2$\,pp on G1 cells, $\pm 3$\,pp on G2 cells, and $\pm 5$--$6$\,pp on G3 cells (smallest tier). Round-3 vs.\ S1 differences on standard tiers are well outside this band on G1/G2 and at the edge on G3; vague-tier differences are well outside on G2 but smaller than the CI on G1 and G3.


\section{Ethics, Risks, and Artifacts}
\label{app:ethics}

\subsection{Upstream Artifacts and Licenses}
\label{app:licenses}

This work builds on the following publicly available artifacts, used in a manner consistent with their stated intended use (research benchmarks and research-grade pretrained models).

\begin{itemize}
\item \textbf{ToolBench} \citep{qin2024toolllm}: source of the underlying API pool and the official G1/G2/G3 evaluation queries. Released under Apache 2.0. \url{https://github.com/OpenBMB/ToolBench}.
\item \textbf{ToolGen} \citep{wang2024toolgen}: source of the 46{,}980-tool catalog from which we derive the 10K subset $\mathcal{T}$, and the source of the (query, gold-tool-set) training pairs $\mathcal{D}_{\mathrm{train}}$. Released under Apache 2.0. \url{https://github.com/Reason-Wang/ToolGen}.
\item \textbf{BGE-large-en-v1.5} \citep{xiao2023bge}: encoder initialisation $\theta_0$. Released under MIT. \url{https://huggingface.co/BAAI/bge-large-en-v1.5}.
\item \textbf{Qwen3.5-4B} \citep{qwen3report}: rewriter initialisation $\psi_0$. Released under Apache 2.0. \url{https://huggingface.co/Qwen/Qwen3.5-4B}.
\end{itemize}

\subsection{Data Coverage and Privacy}
\label{app:data-coverage}

\paragraph{Language and domain.} The ToolBench/ToolGen catalog is entirely English-language and is sourced from RapidAPI's public catalog, skewed toward consumer-facing REST APIs (weather, sports, lifestyle, finance, entertainment). No non-English text appears in queries, tool descriptions, or rewriter outputs.

\paragraph{Personally identifying information.} Tool records contain API metadata (titles, endpoints, parameter schemas, free-text descriptions written by API publishers). They do not contain end-user PII. We did not run a dedicated PII scan on the catalog because the source records are already public API documentation; however, the manual review of 100 vague-paraphrase outputs in Appendix~\ref{app:vague} did not surface any inadvertent generation of personal information.

\paragraph{Offensive content.} The catalog includes some adult-content-tagged APIs (a small minority, consistent with RapidAPI's public listings). We did not filter these out, on the grounds that doing so would change the benchmark composition and make our numbers incomparable with prior work on the same catalog. No offensive content appears in any reported example or figure.

\paragraph{Split sizes.} As reported in \S\ref{sec:method}: catalog $|\mathcal{T}| = 10{,}000$; training set $|\mathcal{D}_{\mathrm{train}}| = 104{,}224$ (G1: 44{,}873; G2: 35{,}402; G3: 23{,}949); evaluation queries 593 / 399 / 100 over G1/G2/G3 (1{,}092 total), with vague paraphrases of the same evaluation queries forming $\mathcal{Q}_{\mathrm{vague}}$ of equal size.

\subsection{Risks}
\label{app:risks}

Tool retrieval is a component of larger tool-using agent systems; improvements in retrieval can amplify both desirable and undesirable downstream agent behaviour, depending on the tools in the catalog and the agent's policy over them. Our experiments are run on the ToolBench-derived ToolGen catalog, which inherits whatever selection biases that catalog has --- consumer-facing REST APIs over enterprise or safety-critical tools, English-language descriptions, no audit of the underlying APIs' content. A rewriter aligned to a specific encoder is, in effect, a steering vector over that encoder's retrieval distribution; the same mechanism that closes catalog-misalignment gaps could in principle be used to bias retrieval toward a chosen subset of tools, and any practitioner reusing this method should be aware that the rewriter's behaviour is encoder-specific. We see no near-term dual-use concern beyond what already applies to any open dense retriever or instruction-tuned LLM.

\subsection{AI Assistant Use}
\label{app:ai-assistants}

For prototyping the codebase and experimentation, as well as for writing and editing of this manuscript, Claude Code with the Opus-4.5 model was used; all technical content, experimental design, claims, and figures are the authors' own.

\end{document}